\documentclass[11pt]{article}

\let\counterwithin\relax

\usepackage[left]{lineno} 

\usepackage{amsmath, amsfonts, amssymb, amsthm, bm, graphicx, mathtools, enumerate,multirow}
\usepackage{natbib}
\usepackage[CJKbookmarks=true,
bookmarksnumbered=true,
bookmarksopen=true,
colorlinks=true,
citecolor=blue,
linkcolor=blue,
anchorcolor=blue,
urlcolor=blue]{hyperref}
\usepackage[usenames]{color}
\usepackage[letterpaper, left=1.2truein, right=1.2truein, top = 1.2truein,
bottom = 1.2truein]{geometry}
\usepackage[ruled, lined, commentsnumbered]{algorithm2e}
\usepackage{prettyref,soul}

\usepackage{apptools} 
\usepackage{chngcntr} 
\AtAppendix{\counterwithin{lemma}{section}} 
\usepackage{tikz}
\usepackage[font=small,labelfont=bf]{caption}
\usepackage{enumitem}

\theoremstyle{definition}

\newrefformat{eq}{(\ref{#1})}
\newrefformat{chap}{Chapter~\ref{#1}}
\newrefformat{sec}{Section~\ref{#1}}
\newrefformat{algo}{Algorithm~\ref{#1}}
\newrefformat{fig}{Fig.~\ref{#1}}
\newrefformat{tab}{Table~\ref{#1}}
\newrefformat{rmk}{Remark~\ref{#1}}
\newrefformat{clm}{Claim~\ref{#1}}
\newrefformat{def}{Definition~\ref{#1}}
\newrefformat{cor}{Corollary~\ref{#1}}
\newrefformat{lmm}{Lemma~\ref{#1}}
\newrefformat{lemma}{Lemma~\ref{#1}}
\newrefformat{prop}{Proposition~\ref{#1}}
\newrefformat{app}{Appendix~\ref{#1}}
\newrefformat{ex}{Example~\ref{#1}}
\newrefformat{cond}{Condition~\ref{#1}}

\usepackage[T1]{fontenc}
\usepackage[utf8]{inputenc}
\usepackage{authblk}
\title{Multi-scale Masked Autoencoder for Electrocardiogram Anomaly Detection}

\author[1,\#,*]{Ya Zhou}
\author[1,\#]{Yujie Yang}
\author[1]{Jianhuang Gan} 
\author[2]{Xiangjie Li}
\author[1]{Jing Yuan}
\author[3,*,\dag]{Wei Zhao }
\affil[1]{Department of Information Center, Fuwai Hospital, Chinese Academy of Medical Sciences and Peking Union Medical College, Beijing, 100037, China}
\affil[2]{
	National Clinical Research Center for Cardiovascular Diseases, Fuwai Hospital, Chinese  Academy of Medical Sciences and Peking Union Medical College, National Center for Cardiovascular Diseases, Beijing, 100037, China 
}
\affil[3]{Fuwai Hospital, National Center for Cardiovascular Diseases, Chinese Academy of Medical Sciences and Peking Union Medical College, Beijing, 100037, China}

\date{}

\begin{document}
\maketitle
\def\thefootnote{\#}\footnotetext{The authors contributed equally}
\def\thefootnote{\arabic{footnote}}
\def\thefootnote{*}\footnotetext{Email: Wei Zhao (zw@fuwai.com), Ya Zhou (zhouya@fuwai.com)
}\def\thefootnote{\arabic{footnote}}
\def\thefootnote{ \dag}\footnotetext{Supervision}

\begin{abstract}	
Electrocardiogram (ECG) analysis is a fundamental tool for diagnosing cardiovascular conditions, yet anomaly detection in ECG signals remains challenging due to their inherent complexity and variability. We propose Multi-scale Masked Autoencoder for ECG anomaly detection (MMAE-ECG), a novel end-to-end framework that effectively captures both global and local dependencies in ECG data. Unlike state-of-the-art methods that rely on heartbeat segmentation or R-peak detection, MMAE-ECG eliminates the need for such pre-processing steps, enhancing its suitability for clinical deployment. MMAE-ECG partitions ECG signals into non-overlapping segments, with each segment assigned learnable positional embeddings. A novel multi-scale masking strategy and multi-scale attention mechanism, along with distinct positional embeddings, enable a lightweight Transformer encoder to effectively capture both local and global dependencies. The masked segments are then reconstructed using a single-layer Transformer block, with an aggregation strategy employed during inference to refine the outputs. Experimental results demonstrate that our method achieves performance comparable to state-of-the-art approaches while significantly reducing computational complexity—approximately 1/78 of the floating-point operations (FLOPs) required for inference. Ablation studies further validate the effectiveness of each component, highlighting the potential of multi-scale masked autoencoders for anomaly detection. 


\end{abstract}

\section{Introduction}
\label{sec:intro}

Electrocardiograms (ECG) are widely used in clinical practice as an affordable and non-invasive tool for diagnosing cardiovascular conditions \citep{somani2021deep}. Early detection of anomalies in ECG signals plays a crucial role in identifying cardiovascular diseases. Traditional automatic ECG analysis algorithms, based on supervised learning, typically require large, labeled datasets for training \citep{ribeiro2020automatic}. While recent advances in self-supervised learning have alleviated the need for extensive labeled data, these methods still require labeled datasets for fine-tuning \citep{zhou2023masked}. Given the diversity and rarity of cardiac diseases \citep{jiang2024self}, such approaches may struggle to detect new abnormal conditions that were not present in the training data. In contrast, anomaly detection, which relies solely on normal healthy data for training, has the potential to identify previously unseen anomalies and mitigate the risk of missing rare cardiac conditions.

Anomaly detection has been widely applied to time-series data analysis, particularly in domains such as economics, manufacturing, and healthcare \citep{zamanzadeh2024deep}. However, its application to ECG signal analysis remains relatively underexplored \citep{jiang2023multi}. 
ECG anomaly detection poses unique challenges 
due to inter-individual variability and inter-sample variability, as well as the complex nature of anomalies, which can manifest in both global rhythm patterns and localized morphological features, further complicating accurate detection and generalization \citep{liu2022time, jiang2023multi}.  
To address these challenges, existing methods such as BeatGAN \citep{liu2022time}, a generative adversarial network designed for heartbeat reconstruction, have demonstrated promise in capturing local morphological features of ECG signals. Similarly, a recently proposed multi-scale approach has achieved state-of-the-art performance on the PTB-XL detection and localization benchmark \citep{jiang2023multi}, highlighting its ability to model both global and local features. Despite these advancements, most current methods rely on R-peak detection or heartbeat segmentation, which introduces additional complexity and makes them highly sensitive to noise and irregularities in the data. This reliance limits their applicability in real-world clinical settings, where reliable R-peak identification may not always be feasible, particularly in noisy or pathological ECG recordings. Consequently, there is a critical need for a new model that can effectively capture both global and local features of ECG signals without depending on R-peak detection, ensuring greater robustness and practicality in clinical applications.

Masked Autoencoders \citep[MAE,][]{he2021masked} have emerged as a powerful self-supervised representation learning technique, excelling in tasks such as image and time-series analysis \citep{he2021masked,zhou2023masked}. While traditional autoencoders are widely regarded as effective for anomaly detection, prior research indicates that MAE may underperform in unsupervised anomaly detection tasks \citep{reiss2022anomaly}. 
The self-attention mechanism in MAE enables it to focus on important input regions, making it adept at capturing global patterns \citep{vaswani2017attention}. 
However, anomaly detection tasks, particularly for ECG signals, require not only capturing global patterns but also identifying subtle, localized features essential for accurate diagnosis.  
These limitations underscore the necessity of adapting MAE to effectively extract both global and local features simultaneously, enabling it to address the specific requirements of ECG anomaly detection.

To overcome these challenges, we propose a novel multi-scale MAE framework for ECG anomaly detection, referred to as MMAE-ECG, which eliminates the need for R-peak detection or heartbeat segmentation. Our approach leverages a Transformer-based encoder-decoder architecture that integrates a novel multi-scale masking strategy, a multi-scale attention mechanism, and distinct positional embeddings to effectively capture both local and global dependencies in ECG signals.
Additionally, an aggregation strategy is employed during inference to refine model predictions.
Evaluations on the PTB-XL anomaly detection and localization benchmark demonstrate that our method matches state-of-the-art performance while requiring only 1/78 of the floating-point operations (FLOPs) for inference, significantly improving computational efficiency. Ablation studies further validate the effectiveness of key components, including multi-scale representation learning, local positional embeddings, multi-scale masking, and the aggregation strategy during inference.

The contributions of this work are summarized as follows: \begin{itemize} 
	\item We propose a novel end-to-end multi-scale MAE framework for both ECG anomaly detection and localization, without the need for R-peak detection.
	\item We introduce a multi-scale masking strategy and multi-scale attention mechanisms to effectively capture both global and local 
	dependencies in ECG signals. 
	\item Experiments on the PTB-XL detection and localization benchmark demonstrate that our method achieves performance comparable to state-of-the-art approaches, while significantly reducing computational complexity—requiring approximately 1/78 of the FLOPs for inference.
	\item We perform ablation studies to evaluate the impact of key design choices, providing valuable insights into the effectiveness of each component in our model.
\end{itemize}

The remainder of the paper is organized as follows: Section II reviews related work in the field of anomaly detection and localization for time-series data. Section III presents the proposed method in detail, followed by the experimental setup and results in Section IV. Section V discusses the results, and Section VI concludes the paper, highlighting potential avenues for future research.

\section{Related work}
\label{sec:related_work}
\subsection{Anomaly Detection in Time Series}
Anomaly detection in time series data has attracted significant attention in recent years due to its diverse applications in domains such as economics, manufacturing, and healthcare \citep{zamanzadeh2024deep}. Existing approaches can be broadly classified into traditional machine learning-based methods \citep{Salem2014AnomalyDI, 10.14778/3467861.3467863, 5437603} and deep learning-based methods \citep{Hundman2018DetectingSA, Zong2018DeepAG, 10.1145/3292500.3330672, tuli2022tranad, xu2022anomaly, zheng2022task, liu2022time}. Deep learning-based methods have demonstrated significant advantages over traditional approaches, achieving superior performance in a variety of real-world time series anomaly detection tasks \citep{zamanzadeh2024deep}. These methods leverage the ability of neural networks to model complex temporal dependencies and capture non-linear patterns inherent in time series data. In this work, we focus on deep learning-based methods.




\subsection{ECG Anomaly Detection}
The diversity and rarity of cardiac diseases, coupled with the high cost of collecting diverse ECG abnormalities, present significant challenges to conventional multi-label classification methods. In contrast, anomaly detection methods, which rely exclusively on normal data for training, offer the potential to identify previously unseen anomalies and reduce the risk of missing rare cardiac conditions. However, ECG anomaly detection remains particularly challenging due to substantial inter-individual and inter-sample variability, as well as the intricate nature of anomalies, which can manifest as both global rhythm disturbances and localized morphological irregularities \citep{liu2022time, jiang2023multi}. To address these issues, generative adversarial network (GAN)-based methods have been explored, such as BeatGAN \citep{liu2022time}, which demonstrates strong capabilities in capturing local morphological features. Similarly, approaches like \cite{qin2023novel} and \cite{wang2023ecggan} leverage GANs to process ECG signals. For jointly modeling local and global ECG patterns,  \cite{jiang2023multi} proposed a multi-scale framework that achieved state-of-the-art results on the PTB-XL detection and localization benchmark \citep{wagner2020ptb, jiang2023multi}. More recently, \cite{bui2024tsrnet} proposed a model that integrates both time-series and time-frequency representations of ECG signals while significantly reducing trainable parameters compared to previous methods. Despite these advancements, most approaches rely on R-peak detection or heartbeat segmentation, which introduces additional complexity and renders them highly sensitive to noise and irregularities, limiting their practicality in real-world clinical settings. To overcome these limitations, our proposed method eliminates the dependence on R-peak detection and heartbeat segmentation. Furthermore, it is lightweight, requiring fewer trainable parameters than current state-of-the-art algorithms, offering significant advantages in efficiency and inference speed.

\subsection{Masked Autoencoders}
In recent years, the focus of deep learning research has shifted from developing increasingly complex models to addressing the challenges of data scarcity \citep{zhang2023survey}. Masked Autoencoders \citep[MAE,][]{he2021masked} have emerged as a powerful self-supervised representation learning framework, demonstrating remarkable success in various visual tasks and gaining significant attention. Recently, efforts have been made to adapt MAE for ECG classification \citep{zhang2022maefe, yang2022masked, sawano2022masked, wang2023unsupervised, zhou2023masked}. Among these, \cite{zhou2023masked} proposed an MAE-based multi-label ECG classification approach, achieving notable performance improvements. However, the application of MAE to anomaly detection remains limited. For instance,  \cite{reiss2022anomaly} observed that MAE may underperform in unsupervised anomaly detection tasks for images. While the self-attention mechanism inherent in MAE enables it to capture global patterns effectively \citep{vaswani2017attention}, anomaly detection tasks, particularly for ECG signals, require not only modeling global features but also detecting subtle and localized anomalies critical for accurate diagnosis. To address these challenges, we propose a novel multi-scale MAE-based framework specifically designed for ECG anomaly detection.

\section{Methodology}
\label{sec:method}
\begin{figure*}[!hbt]
	\centering
	\includegraphics[width=1\linewidth]{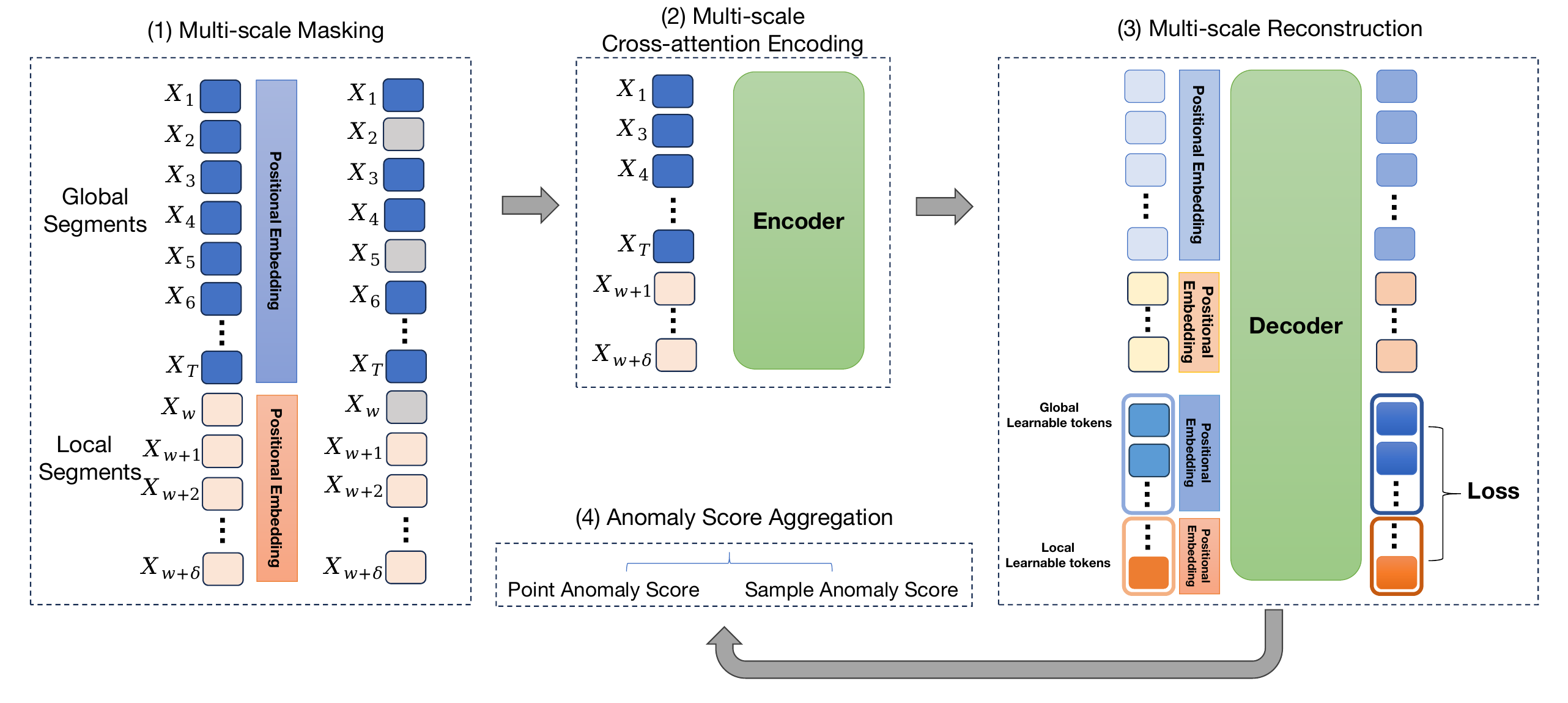}
	\caption{Overview of the proposed framework. (1) Multi-scale Masking: Segments in the global and local regions are masked separately. (2) Multi-scale Cross-attention Encoding: Unmasked segments from both regions are concatenated and fed into a lightweight Transformer-based encoder for cross-attention. (3) Multi-scale Reconstruction: Masked segments in global and local regions are reconstructed using a single-layer Transformer block based on mean square loss after per-segment normalization. (4) Anomaly Score Aggregation: An aggregation strategy enhances sample-level and point-level anomaly scores for anomaly detection and localization, respectively.
		\label{frameworks}}
\end{figure*}

Our proposed framework consists of four key components: (1) multi-scale masking, (2) multi-scale cross-attention encoding, (3) multi-scale reconstruction, and (4) anomaly score aggregation. An overview of the framework is illustrated in Figure \ref{frameworks}. In the following, we provide a detailed explanation of each component.

\subsection{Multi-scale Masking}
Let a multi-lead ECG signal be denoted as \( \mathbf{X}^\sharp = (X_{k,q}^\sharp) \in \mathbb{R}^{K \times Q} \), where \( K \) represents the number of leads, and \( Q \) is the length of the ECG signal. Following \cite{zhou2023masked}, we partition the ECG signal \( \mathbf{X}^\sharp \) along the time dimension into a sequence of non-overlapping segments as follows:
\begin{equation*}
	\mathcal{U} := \{\mathbf{X}_1, \cdots, \mathbf{X}_T\},
\end{equation*}
where \( T \) is the total number of segments. Each segment \( \mathbf{X}_t = (X_{k,q}^t) \in \mathbb{R}^{K \times (Q/T)} \) represents a subset of the original signal, with \( X_{k,q}^t = X_{k, \{(t-1)\cdot Q/T + q\}}^\sharp \) for \( t = 1, \ldots, T \), \( k = 1, \ldots, K \), and \( q = 1, \ldots, Q/T \).

Next, we select multiple consecutive segments from \( \mathcal{U} \) to construct a sequence of local regions \( \mathcal{V}^{w_1}, \cdots, \mathcal{V}^{w_\nu} \), where each local region is defined as:
\begin{equation*}
	\mathcal{V}^w := \{\mathbf{X}_{w+1}, \cdots, \mathbf{X}_{w+\delta}\},
\end{equation*}
for \( w = w_1, \ldots, w_\nu \) and \( 0 \leq w_1 < w_2 < \ldots < w_\nu \leq T - \delta \), where \( \delta \) is the predefined length of the local region.

During training, for each batch, we randomly select \( w \in \{w_1, \ldots, w_\nu\} \) and separately apply masking to the elements in \( \mathcal{U} \) and \( \mathcal{V}^w \). Specifically, given a masking ratio \( \theta \), we uniformly sample \( S := \min\{\max\{[T\theta], 1\}, T-1\} \) segments from \( \mathcal{U} \), and \( R := \min\{\max\{[\delta \theta], 1\}, \delta - 1\} \) segments from \( \mathcal{V}^w \), which are then masked. For notational simplicity, we denote the masked segments as:
\[
\mathcal{U}_{mask} = \{\mathbf{X}_{j_1}, \cdots, \mathbf{X}_{j_{S^\prime}}\}
\]
and
\[
\mathcal{V}_{mask}^w = \{\mathbf{X}_{j_{w,1}}, \cdots, \mathbf{X}_{j_{w, R^\prime}}\},
\]
where \( j_s, s = 1, \ldots, S^\prime \) and \( j_{w,r}, r = 1, \ldots, R^\prime \) are randomly chosen from the index sets \( \{1, \ldots, T\} \) and \( \{w+1, \ldots, w+\delta\} \), respectively.

Similarly, the unmasked segments are denoted as:
\[
\mathcal{U}_{unmask} = \{\mathbf{X}_{i_1}, \cdots, \mathbf{X}_{i_{S}}\}
\]
and
\[
\mathcal{V}_{unmask}^w = \{\mathbf{X}_{i_{w,1}}, \cdots, \mathbf{X}_{i_{w,R}}\},
\]
where \( i_s, s = 1, \ldots, S \) and \( i_{w,r}, r = 1, \ldots, R \) represent the indices of the unmasked segments. Using these notations, we can express the total set of segments as:
\[
\mathcal{U} = \mathcal{U}_{unmask} \cup \mathcal{U}_{mask}, \quad \text{with} \quad \mathcal{U}_{unmask} \cap \mathcal{U}_{mask} = \varnothing,
\]
and
\[
\mathcal{V}^w = \mathcal{V}_{unmask}^w \cup \mathcal{V}^w_{mask}, \quad \text{with} \quad \mathcal{V}^w_{unmask} \cap \mathcal{V}^w_{mask} = \varnothing.
\]
Here, \( \mathcal{U}_{unmask} \) and \( \mathcal{V}_{unmask}^w \) are fed into the encoder to achieve multi-scale cross-attention, while \( \mathcal{U}_{mask} \) and \( \mathcal{V}_{mask}^w \) serve as the reconstruction targets.

\subsection{Multi-scale  Cross-attention Encoding}

We introduce a self-attention mechanism to model the relationships between global and local features. To achieve this, we first concatenate the unmasked elements from \( \mathcal{U}_{unmask} \) and \( \mathcal{V}_{unmask} \). To preserve sequence order information, we adopt the approach in \cite{zhou2023masked}, using learnable positional embeddings. However, applying standard positional embeddings without distinguishing between local and global features could lead to the model overlooking their positional differences. To address this, we introduce distinct positional embeddings for local and global features, enabling the model to better capture and differentiate the unique characteristics of each feature set.

We now describe the encoding module in detail. Denote the layer normalization \citep{ba2016layer}, multi-headed self-attention, and multi-layer perceptron (MLP) blocks, as introduced in  \cite{dosovitskiy2020image}, by \( LN(\cdot) \), \( MSA(\cdot) \), and \( MLP(\cdot) \), respectively. For simplicity, let \( \mathbf{x}_{i_s}^\top \) and \( \mathbf{x}_{i_{w,r}}^\top \in \mathbb{R}^{KQ/T} \) represent the vectorized forms of \( \mathbf{X}_{i_s} \) for \( s = 1, \ldots, S \) and \( \mathbf{X}_{i_{w,r}} \) for \( r = 1, \ldots, R \). Let \( D \) denote the latent vector size. Define the linear projection matrix \( \mathbf{E} \in \mathbb{R}^{(KQ/T) \times D} \), the auxiliary token \( \mathbf{x}_{aux}^\top \in \mathbb{R}^D \), and the learnable positional embedding vector \( \mathbf{e}_{pos} = (\mathbf{e}_0, \mathbf{e}_1, \cdots, \mathbf{e}_T, \mathbf{e}_{T+1}, \ldots, \mathbf{e}_{T+\delta})^\top \in \mathbb{R}^{D(T+\delta+1)} \). Here, \( \mathbf{e}_t^\top \in \mathbb{R}^D, t = 0, 1, \ldots, T \) are used to preserve the sequential order information for global features, while \( \mathbf{e}_{T+t}^\top \in \mathbb{R}^D, t = 1, \ldots, \delta \) are employed to encode local features.

Only the unmasked segments from \( \mathcal{U} \) and \( \mathcal{V}^w \) are passed through the model. The input representation is defined as:
\begin{equation*}
	\mathbf{z}_0 = [\mathbf{x}_{aux}; \mathbf{x}_{i_1} \mathbf{E}; \cdots; \mathbf{x}_{i_S} \mathbf{E}; \mathbf{x}_{i_{w,1}} \mathbf{E}; \cdots; \mathbf{x}_{i_{w,R}} \mathbf{E}] + [\mathbf{e}_0; \mathbf{e}_{i_1}; \cdots; \mathbf{e}_{i_S}; \mathbf{e}_{T+i_{w,1}}; \cdots; \mathbf{e}_{T+i_{w,R}}],
\end{equation*}
where \( \mathbf{x}_{i_s} \mathbf{E} \) and \( \mathbf{e}_{i_s} \) denote the projections of the unmasked global segments and their corresponding positional embeddings, and \( \mathbf{x}_{i_{w,r}} \mathbf{E} \) and \( \mathbf{e}_{T+i_{w,r}} \) represent the projections of the unmasked local segments along with their respective embeddings.

The encoding process consists of multiple layers of self-attention and MLP blocks:
\begin{equation*}
	\mathbf{z}_l^\prime = MSA(LN(\mathbf{z}_{l-1})) + \mathbf{z}_{l-1}, \quad l = 1, \ldots, L,
\end{equation*}
\begin{equation*}
	\mathbf{z}_l = MLP(LN(\mathbf{z}_l^\prime)) + \mathbf{z}_l^\prime, \quad l = 1, \ldots, L,
\end{equation*}
where \( L \) denotes the number of transformer blocks.

Finally, the output of the encoder is given by:
\begin{equation*}
	\mathbf{z}_L = [\mathbf{z}_L^0; \mathbf{z}_{i_1}^L; \cdots; \mathbf{z}_{i_S}^L; \mathbf{z}_{i_{w,1}}^L; \cdots; \mathbf{z}_{i_{w,R}}^L],
\end{equation*}
where \( \mathbf{z}_L^0 \in \mathbb{R}^D \) represents the encoded auxiliary token, \( \mathbf{z}_{i_s}^L \in \mathbb{R}^D \) are the encoded unmasked global segments, and \( \mathbf{z}_{i_{w,r}}^L \in \mathbb{R}^D \) are the encoded unmasked local segments. These encoded representations are subsequently used to reconstruct the global and local features, respectively.

\subsection{Multi-Scale Reconstruction}

In this section, we present the multi-scale reconstruction strategy that employs a Transformer-based decoder. This decoder helps encourage the encoder to learn meaningful wave shape features. Specifically, we adopt a one-layer Transformer decoder. Let \( D^\prime \) denote the latent vector size. We define the learnable components as follows: \( \mathbf{E}^\prime \in \mathbb{R}^{D \times D^\prime} \), \( \mathbf{E}_0 \in \mathbb{R}^{D^\prime \times (KQ/T)} \), \( \mathbf{e}_m^\top \in \mathbb{R}^{D^\prime} \), and the positional embeddings \( \mathbf{e}_{pos}^\prime = (\mathbf{e}_1^\prime, \cdots, \mathbf{e}_T^\prime, \mathbf{e}_{T+1}^\prime, \cdots, \mathbf{e}_{T+\delta}^\prime)^\top \in \mathbb{R}^{(T + \delta) D^\prime} \). Here, \( \mathbf{e}_t^\prime \in \mathbb{R}^{D^\prime} \) for \( t = 1, \ldots, T \) corresponds to the positional embeddings for global features, and \( \mathbf{e}_{T+t}^\prime \in \mathbb{R}^{D^\prime} \) for \( t = 1, \ldots, \delta \) serves as the positional embeddings for local features. Additionally, \( \mathbf{e}_m^\top \) represents the embeddings for the masked segments.

The decoder can be formulated as follows:
\begin{equation*}
	\label{def:decoder}
	\begin{aligned}
		\tilde{\mathbf{z}}_0 &= [\mathbf{z}_L^{i_1} \mathbf{E}^\prime; \cdots; \mathbf{z}_L^{i_S} \mathbf{E}^\prime; \mathbf{z}_L^{i_{w,1}} \mathbf{E}^\prime; \cdots; \mathbf{z}_L^{i_{w,R}} \mathbf{E}^\prime; \mathbf{e}_m; \cdots; \mathbf{e}_m] \\
		& \quad \quad + [\mathbf{e}_{i_1}^\prime; \cdots; \mathbf{e}_{i_S}^\prime; \mathbf{e}_{i_{w,1}}^\prime; \cdots; \mathbf{e}_{i_{w,R}}^\prime; \mathbf{e}_{j_1}^\prime; \cdots; \mathbf{e}_{j_{S^\prime}}^\prime; \mathbf{e}_{j_{w,1}}^\prime; \cdots; \mathbf{e}_{j_{w,S^\prime}}^\prime],
		\\
		\tilde{\mathbf{z}}_1^\prime &= MSA(LN(\tilde{\mathbf{z}}_0)) + \tilde{\mathbf{z}}_0, \\
		\tilde{\mathbf{z}}_1 &= MLP(LN(\tilde{\mathbf{z}}_1^\prime)) + \tilde{\mathbf{z}}_1^\prime.
	\end{aligned}
\end{equation*}

Here, \( \tilde{\mathbf{z}}_1^\prime \) is given by
\[
\tilde{\mathbf{z}}_1^\prime = [\tilde{\mathbf{z}}_1^{i_1}; \cdots; \tilde{\mathbf{z}}_1^{i_S}; \tilde{\mathbf{z}}_1^{i_{w,1}}; \cdots; \tilde{\mathbf{z}}_1^{i_{w,R}}; \tilde{\mathbf{z}}_1^{j_1}; \cdots; \tilde{\mathbf{z}}_1^{j_{S^\prime}}; \tilde{\mathbf{z}}_1^{j_{w,1}}; \cdots; \tilde{\mathbf{z}}_1^{j_{w,R^\prime}}],
\]
where each \( (\tilde{\mathbf{z}}_1^{i_s})^\top, (\tilde{\mathbf{z}}_1^{j_{s^\prime}})^\top, (\tilde{\mathbf{z}}_1^{i_{w,r}})^\top, (\tilde{\mathbf{z}}_1^{j_{w,r^\prime}})^\top \in \mathbb{R}^{D^\prime} \) for \( s = 1, \ldots, S \), \( s^\prime = 1, \ldots, S^\prime \), \( r = 1, \ldots, R \), and \( r^\prime = 1, \ldots, R^\prime \). The segments \( \tilde{\mathbf{z}}_1^{j_{s^\prime}} \) and \( \tilde{\mathbf{z}}_1^{j_{w,r^\prime}} \) are used to reconstruct the global and local masked segments, respectively.

The decoder outputs are obtained by:
\begin{equation*}
	\label{def:decoder_output}
	\tilde{\mathbf{x}}_{j_{s^\prime}} = \tilde{\mathbf{z}}_1^{j_{s^\prime}} \mathbf{E}_0, \quad s^\prime = 1, \ldots, S^\prime,
\end{equation*}
and
\begin{equation*}
	\label{def:decoder_output2}
	\tilde{\mathbf{x}}_{j_{w,r^\prime}} = \tilde{\mathbf{z}}_1^{j_{w,r^\prime}} \mathbf{E}_0, \quad r^\prime = 1, \ldots, R^\prime.
\end{equation*}

During training, the objective is to reconstruct the normalized values of the masked global and local segments. We define the reconstruction loss for the global and local features as:
\[
l_{\text{global}} = \sum_{s^\prime = 1}^{S^\prime} \left\| \tilde{\mathbf{x}}_{j_{s^\prime}} - f(\mathbf{x}_{j_{s^\prime}}) \right\|_2^2,
\]
and
\[
l_{\text{local}}^w = \sum_{r^\prime = 1}^{R^\prime} \left\| \tilde{\mathbf{x}}_{j_{w,r^\prime}} - f(\mathbf{x}_{j_{w,r^\prime}}) \right\|_2^2,
\]
where \( \mathbf{x}_{j_{s^\prime}} \) and \( \mathbf{x}_{j_{w,r^\prime}} \in \mathbb{R}^{QW/T} \) are the vectorized forms of the global and local segments \( \mathbf{X}_{j_{s^\prime}} \) and \( \mathbf{X}_{j_{w,r^\prime}} \), and \( f: \mathbb{R}^{QW/T} \to \mathbb{R}^{QW/T} \) is a predefined per-segment normalization function as specified in \cite{zhou2023masked}. The final loss function is then the sum of the global and local reconstruction losses:
\[
l^w = l_{\text{global}} + l_{\text{local}}^w.
\]

\subsection{Anomaly Score Aggregation}
In the anomaly detection framework, each test sample \( \mathbf{X}^\sharp \) undergoes a sequence of forward passes, where the masking segments are determined randomly in each pass. To ensure that segments within the local region are reconstructed with high probability, we evaluate the test sample through \( H \) independent forward passes. Here, \( H \) is a predefined constant, which ensures that a segment is masked with the probability:
\[
1 - \left( \frac{R}{\delta} \right)^H,
\]
where \( R \) represents the number of masked segments and \( \delta \) is the total number of segments.

To further improve reconstruction accuracy, we leverage multi-scale cross-attention to cover all local regions, including \( \mathcal{V}^{w_1}, \cdots, \mathcal{V}^{w_v} \). For each local region \( \mathcal{V}^{w_i} \) and each forward pass \( h \), we denote the corresponding reconstruction loss as \( l_{local}^{w_i,h} \), for \( i = 1, \ldots, v \) and \( h = 1, \ldots, H \). Additionally, since the global features may also vary across different passes and regions, we use \( l_{global}^{w_i,h} \) to denote the loss associated with the global features for the same \( i \) and \( h \).

The anomaly score for the test sample \( \mathbf{X}^\sharp \) is then defined as the average of the losses across all local regions and forward passes:
\begin{equation}
	\label{def:anomaly_score_sample_level}
	\mathcal{A}(\mathbf{X}^\sharp) := \frac{1}{Hv} \sum_{h=1}^H \sum_{i=1}^v \left( l_{global}^{w_i,h} + l_{local}^{w_i,h} \right).
\end{equation}

For localization of anomalies, the anomaly score for a specific signal point, denoted as \( X_{k,q}^\sharp \), corresponds to the part of the anomaly score in \eqref{def:anomaly_score_sample_level} that is related to that signal point. Specifically, the global and local loss terms \( l_{global}^{w_i,h} \) and \( l_{local}^{w_i,h} \) are aggregated over a subset of signal points. By summing the contributions related to \( X_{k,q}^\sharp \), we define the localized anomaly score \( l_{k,q}^{w_i,h} \) for each forward pass \( h \) and local region \( w_i \). The final anomaly score for the signal point \( X_{k,q}^\sharp \) is given by:
\begin{equation*}
	\label{def:anomaly_score_point_level}
	\mathcal{A}_{k,q}(\mathbf{X}^\sharp) := \frac{1}{Hv} \sum_{h=1}^H \sum_{i=1}^v l_{k,q}^{w_i,h}.
\end{equation*}

\section{Experiments}
\label{sec:exp}
This section presents an evaluation of the proposed method using the PTB-XL anomaly detection and localization benchmark \citep{jiang2023multi}, which offers a comprehensive tool for ECG-based anomaly detection tasks. The dataset contains 10,327 12-lead ECG recordings, sampled at 500 Hz over 10 seconds. The training set consists of 8,167 normal recordings, while the test set includes 912 normal and 1,248 abnormal recordings, covering a wide range of cardiovascular conditions \citep{wagner2020ptb}. For anomaly localization, the dataset provides point-level annotations for 400 ECG recordings across 22 abnormality types, labeled by cardiologists for accuracy.

\subsection{Implementation Details}

In our experiments, we use a segment size of 125, resulting in a sequence length of $T = 40$. We set $\delta = 4$ and define the local regions at the points 1, 5, 9, 13, 17, 21, 25, 29, 33, excluding the segments at the beginning and end of the sequence, similar to \cite{jiang2023multi}. 
The masking ratio is set to $\theta = 25\%$, and the encoder consists of $L = 3$ layers with 16 self-attention heads and a latent dimension of $D = 64$. The decoder has a latent dimension of $D^\prime = 64$ with 2 self-attention heads.  Training uses the AdamW optimizer with a cosine annealing learning rate schedule and a batch size of 256, running for 300 epochs with a warm-up of 40 epochs. For inference, we select $H = 4$ to ensure that each segment in the local regions is masked with at least 99\% probability. Performance is evaluated using the Area Under the Receiver Operating Characteristic Curve (AUC), following \cite{jiang2023multi} and \cite{bui2024tsrnet}.

\subsection{Comparisons with State-of-the-Arts}
We compare our proposed method with several state-of-the-art time-series anomaly detection approaches, including TranAD \citep{tuli2022tranad}, AnoTran \citep{xu2022anomaly}, TSL \citep{zheng2022task}, BeatGAN \citep{liu2022time}, MCF \citep{jiang2023multi} and TSRNet \citep{bui2024tsrnet}. The results of TranAD, AnoTran,  TSL and  MCF are excerpted from \cite{jiang2023multi}, while that of TSRNet is excerpted from \cite{bui2024tsrnet}. 
As shown in Table \ref{tab:methods_comparison}, both MCF and our method significantly outperform baseline models in anomaly detection and localization, with our method achieving comparable detection performance and slightly better localization accuracy. This demonstrates our method's ability to effectively capture both global and local features of ECG signals, offering improved robustness and precision over existing solutions.

Table \ref{tab:comparison_to_MCF} further highlights the computational efficiency of our method. Unlike MCF, which requires R-peak detection during preprocessing, our method eliminates this step, simplifying data preparation. In terms of computational complexity (GFLOPs), MCF requires 45.108 GFLOPs per inference, computed as $1.253 \times 12 \times 3$, where $1.253$ represents the GFLOPs per forward pass, $12$ corresponds to the number of R-peaks, and $3$ accounts for feed-forward operations. Specifically, MCF performs approximately 36 forward passes, based on the median number of R-peaks detected by its implementation \citep{jiang2023multi}, with each pass requiring 1.253 GFLOPs. In contrast, our method requires only 0.576 GFLOPs per inference, computed as $0.016 \times 9 \times 4$, where $0.016$ denotes the GFLOPs per forward pass, $9$ represents the number of local regions, and $4$ corresponds to the number of aggregation operations ($H$). This results in an approximately 78× reduction in computational complexity compared to MCF (0.576 GFLOPs vs. 45.108 GFLOPs), significantly lowering resource demands. Moreover, our approach features a substantially smaller model size (0.398M parameters vs. 7.086M) and a dramatically faster training time (0.225 hours vs. 9.537 hours). The reported training time is the median of five independent runs, conducted on a server equipped with an NVIDIA Tesla V100 GPU and an Intel Xeon Gold 6130 CPU. These improvements in computational efficiency and model complexity make our method particularly well-suited for deployment in resource-constrained environments, enhancing its practical applicability in real-world clinical settings.

\begin{table}[ht]
	\centering
	\caption{Comparison of Methods}
	\begin{tabular}{|c|c|c|c|c|c|c|}
		\hline
		Method   &Detection & Localization \\
		\hline
		TranAD \citep{tuli2022tranad}   &0.788 & 0.685 \\
		AnoTran \citep{xu2022anomaly}  &0.762 & 0.641 \\
		TSL \citep{zheng2022task} &0.757 & 0.509 \\
		BeatGAN \citep{liu2022time} & 0.799 & 0.715 \\
		TSRNet \citep{bui2024tsrnet}  & \textbf{0.860} & - \\
		MCF  \citep{jiang2023multi} & \textbf{0.860} & 0.747 \\
		MMAE-ECG  &\textbf{0.860} & \textbf{0.749} \\
		\hline
	\end{tabular}
	\label{tab:methods_comparison}
\end{table}

\begin{table}[htbp]
	\centering
	\caption{Comparison of Computational Complexity and Model Requirements}
	\begin{tabular}{|l|c|c|}
		\hline
		\textbf{Metric}         & MCF     & MMAE-ECG   \\ 
		\hline
		R-peak Detection & Required & \textbf{Not required}  \\
		\hline
		Inference Complexity (GFLOPs) &  $\approx$ 45.108   & \textbf{0.576}     \\ 
		\hline
		Parameters (M) & 7.086      & \textbf{0.398}                    \\ \hline
		Model Type & Convolutional-based     & Transformer-based                \\ \hline
		Training Time (h) &  9.537                  & \textbf{0.225}                     \\ \hline
	\end{tabular}
	\label{tab:comparison_to_MCF}
\end{table}

\subsection{Visualization for Anomaly Localization}
\begin{figure*}[!hbt]
	\centering
	\begin{minipage}{0.48\linewidth}
		\centering
		\includegraphics[width=\linewidth]{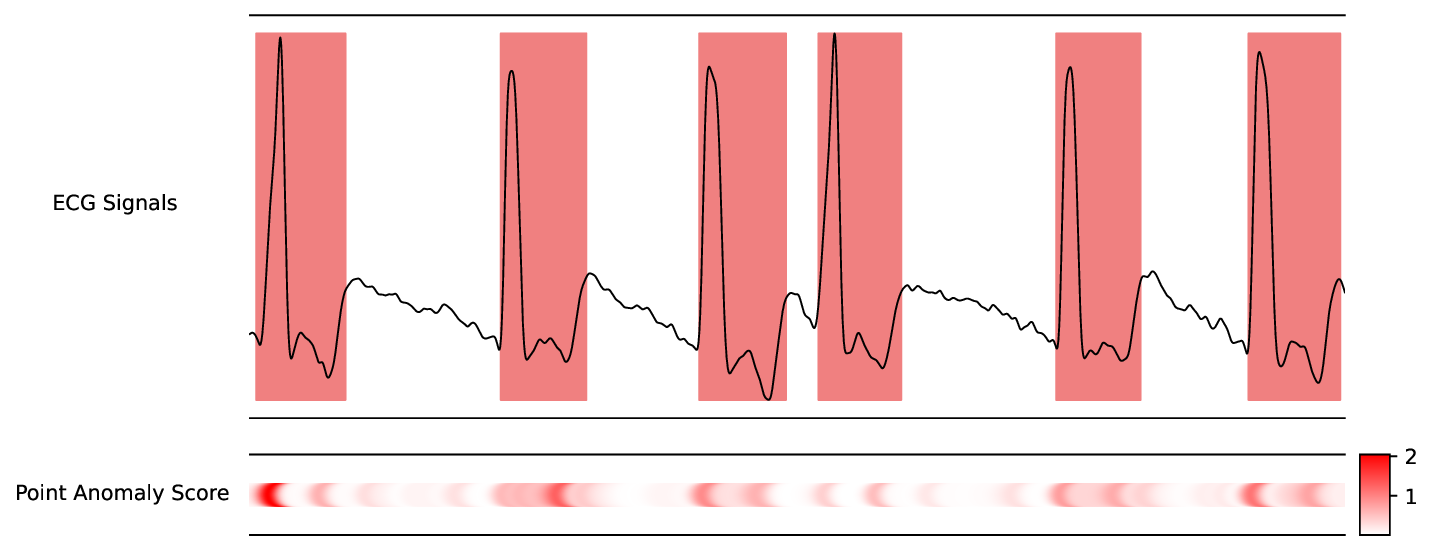} \\ (A)
	\end{minipage}
	\hfill
	\begin{minipage}{0.48\linewidth}
		\centering
		\includegraphics[width=\linewidth]{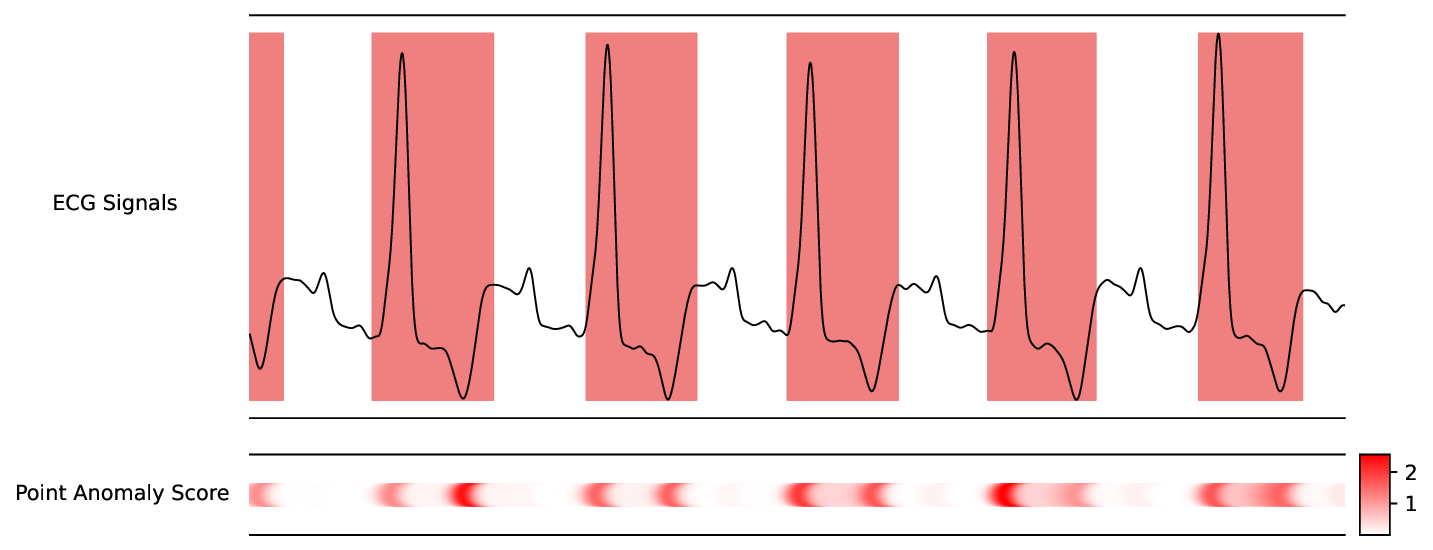} \\ (B)
	\end{minipage}
	
	\begin{minipage}{0.48\linewidth}
		\centering
		\includegraphics[width=\linewidth]{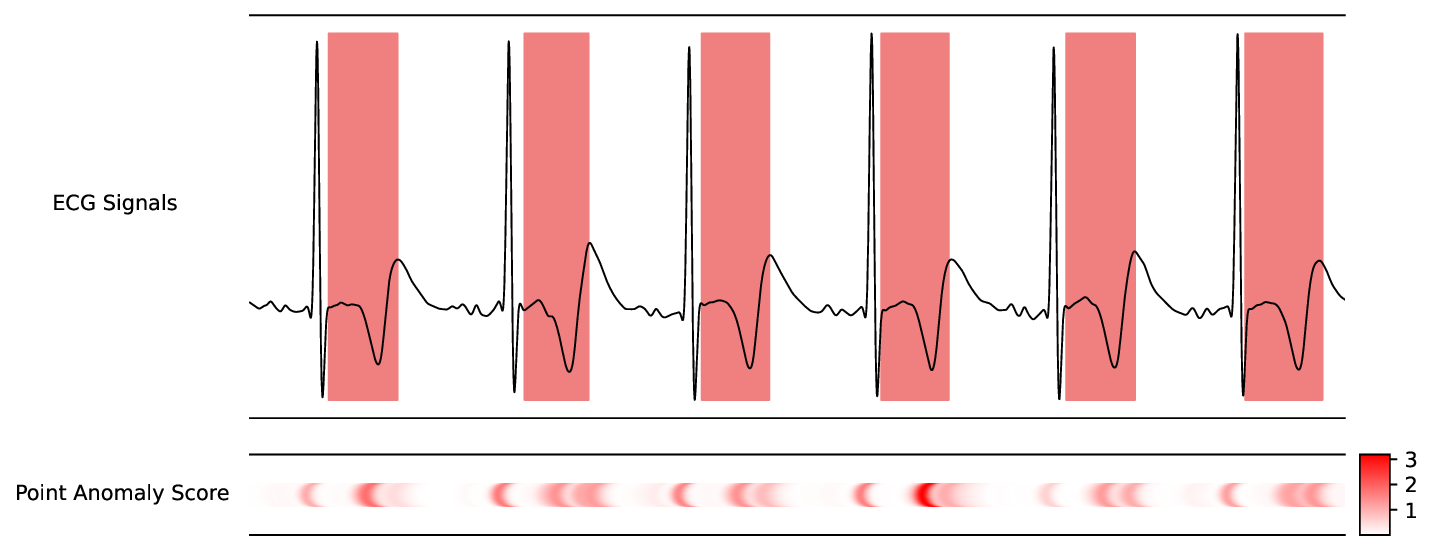} \\ (C)
	\end{minipage}
	\hfill
	\begin{minipage}{0.48\linewidth}
		\centering
		\includegraphics[width=\linewidth]{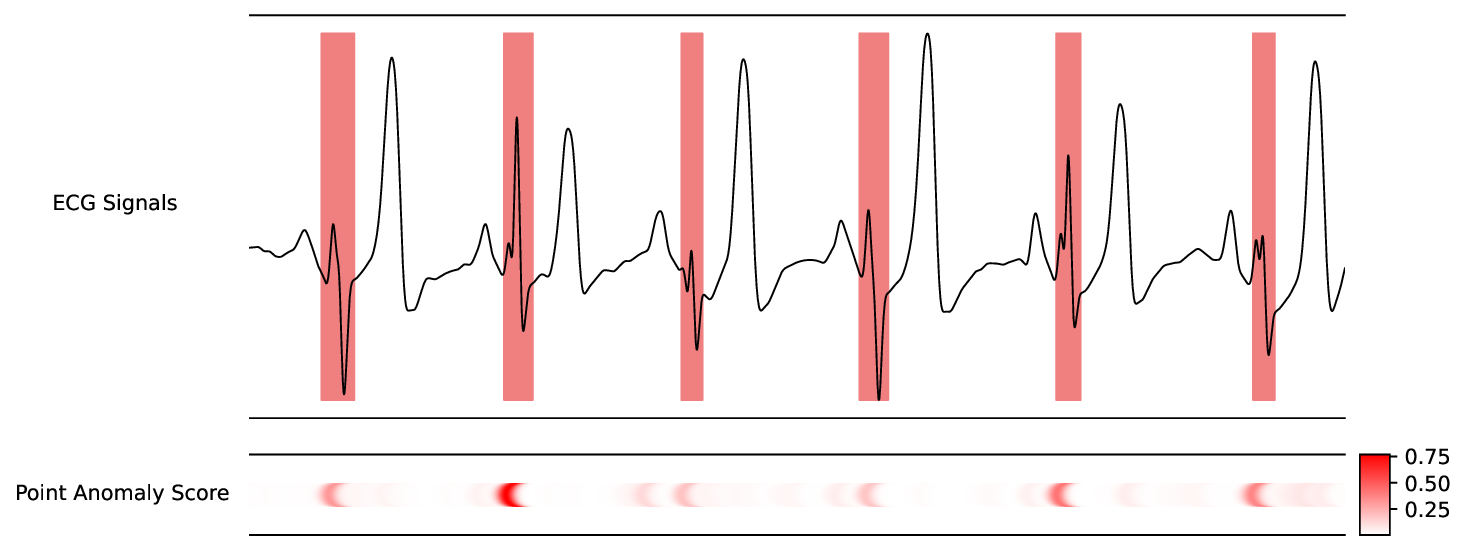} \\ (D)
	\end{minipage}

	\begin{minipage}{0.48\linewidth}
		\centering
		\includegraphics[width=\linewidth]{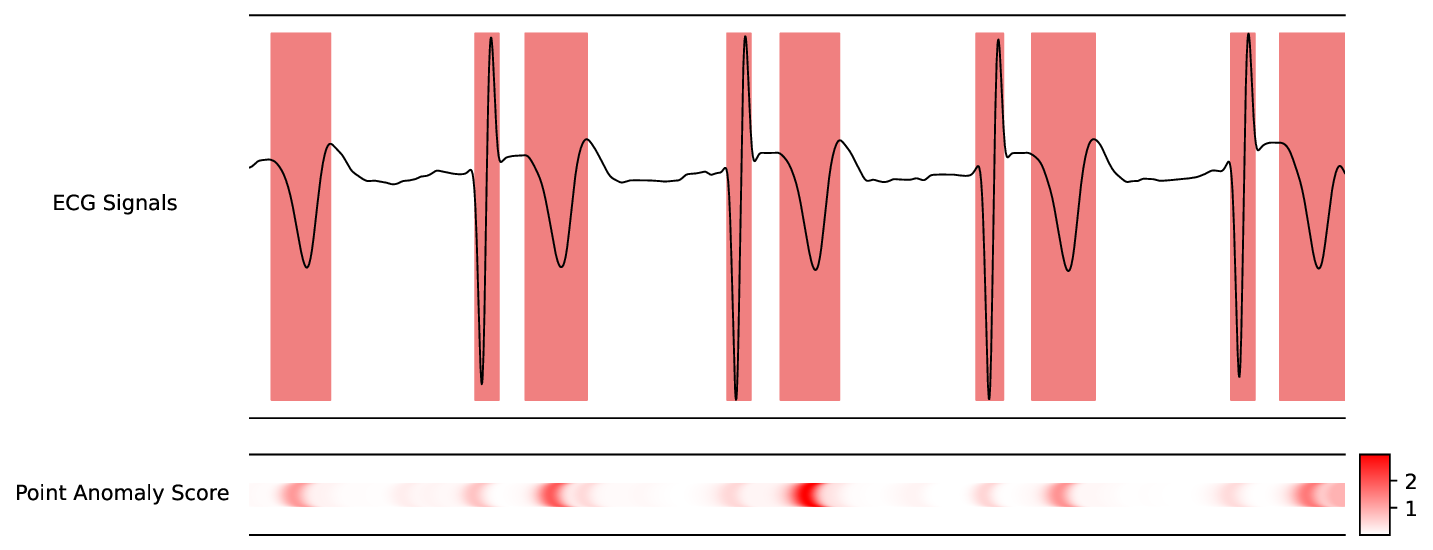} \\ (E)
	\end{minipage}
	\hfill
	\begin{minipage}{0.48\linewidth}
		\centering
		\includegraphics[width=\linewidth]{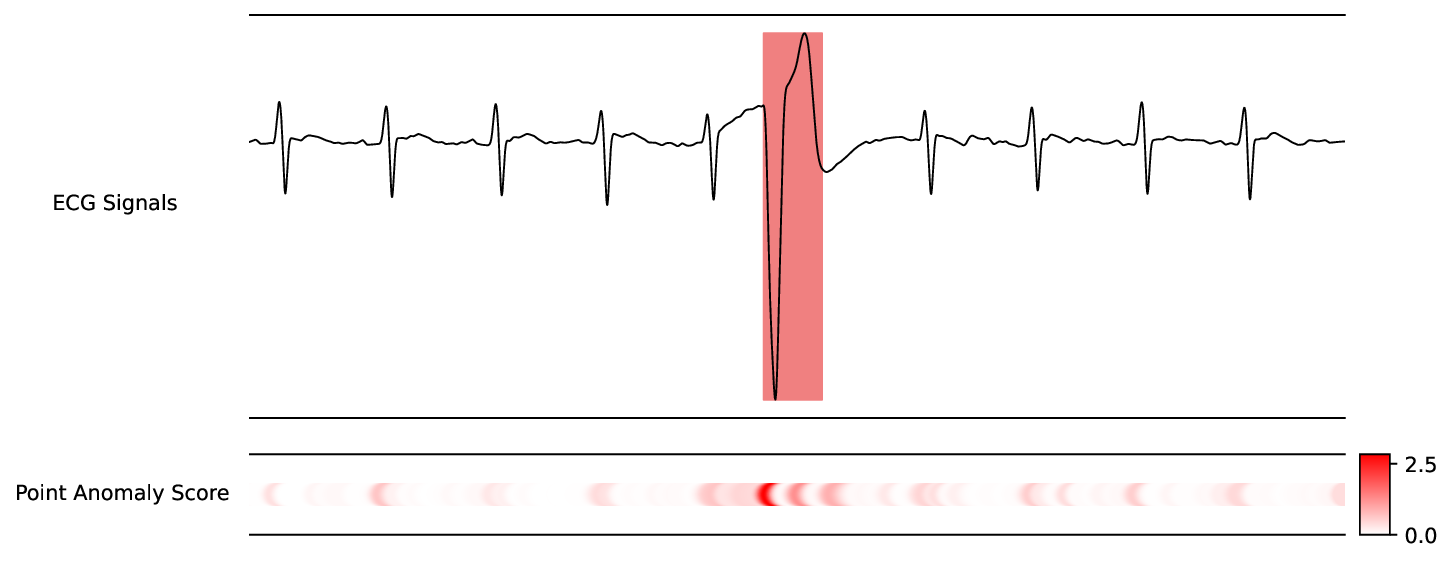} \\ (F)
	\end{minipage}
	
	\caption{Examples of anomaly localization on PTB-XL with different abnormal types. Ground truths are highlighted in red boxes on the ECG signals, and anomaly localization of the proposed method are attached below.}
	\label{fig_anomaly_localization}
\end{figure*}

To further demonstrate the effectiveness of MMAE-ECG in anomaly localization, we present visualization results on representative samples from the PTB-XL benchmark, as shown in Figure \ref{fig_anomaly_localization}. These examples cover a diverse range of ECG abnormalities, as annotated by experienced cardiologists \citep{jiang2023multi}, with detailed descriptions provided in Appendix \ref{app:detail_exam}.
As illustrated in Figure \ref{fig_anomaly_localization}, the proposed method effectively identifies diverse anomaly regions across different ECG leads. These visualizations provide valuable insights into the model’s decision-making process and highlight its potential utility in assisting clinicians with rapid and accurate anomaly localization in real-world applications.

\subsection{Ablation Study}
We conduct ablation studies to systematically evaluate the contribution of each design choice in our model, using the PTB-XL anomaly detection benchmark, which includes patients with diverse characteristics. Specifically, we investigate the following key aspects:

\begin{itemize}
	\item [\textbf{a.}] The impact of multi-scale region utilization. 
	\item [\textbf{b.}] The effectiveness of the local positional embedding.
	\item [\textbf{c.}] The influence of the multi-scale masking strategy.
	\item [\textbf{d.}] The necessity of the masked segment-based loss function.
	\item [\textbf{e.}] The effect of varying masking ratios.
	\item [\textbf{f.}] The influence of different aggregation strategies during inference.
\end{itemize}

We design a series of experiments to evaluate these aspects. The results for experiments \textbf{a} to \textbf{d} are summarized in Table \ref{tab:ablation_study_abcd}. Specifically:
\begin{itemize}
	\item For \textbf{a}, we evaluate the model’s performance by removing either the local region or the global region in our framework.
	\item For \textbf{b}, we replace our specially-designed local positional embedding with the corresponding positional embedding used for global region $\mathcal{U}$.
	\item For \textbf{c}, we replace the multi-scale masking strategy with a single masking approach, where several segments are randomly masked from the concatenated global and local features, potentially leaving all the local segments unmasked.
	\item For \textbf{d}, we modify the loss function to compute the loss over all segments, rather than just the masked segments.
\end{itemize}

The results of these experiments show a significant degradation in anomaly detection performance when the proposed settings are not applied, as detailed in Table \ref{tab:ablation_study_abcd}.

\begin{table}[h]
	\centering
	\begin{tabular}{|c|c|}
		\hline
		\textbf{Configuration} & {AUC} \\
		\hline
		MMAE-ECG & \textbf{0.860} \\
		\textbf{(a)} Global region only & 0.825 \\
		\textbf{(a)} Local region only & 0.793 \\
		\textbf{(b)} Global positional embedding applied to local region & 0.847 \\
		\textbf{(c)} Single mask (vs. Multi-scale mask) & 0.845 \\
		\textbf{(d)} Loss computed on all segments & 0.731 \\
		\hline
	\end{tabular}
	\caption{Ablation study results for different model configurations.}
	\label{tab:ablation_study_abcd}
\end{table}

\begin{figure*}[!hbt]
	\centering
	\begin{minipage}{0.45\linewidth}
		\centering
		\includegraphics[width=1\linewidth]{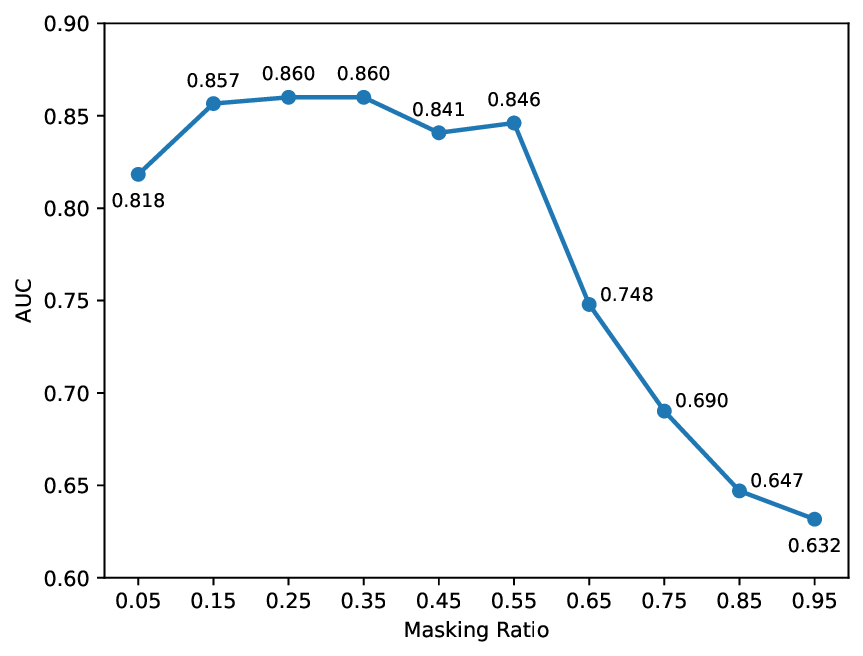} \\ (A)
	\end{minipage}
	\hfill
	\begin{minipage}{0.45\linewidth}
		\centering
		\includegraphics[width=1\linewidth]{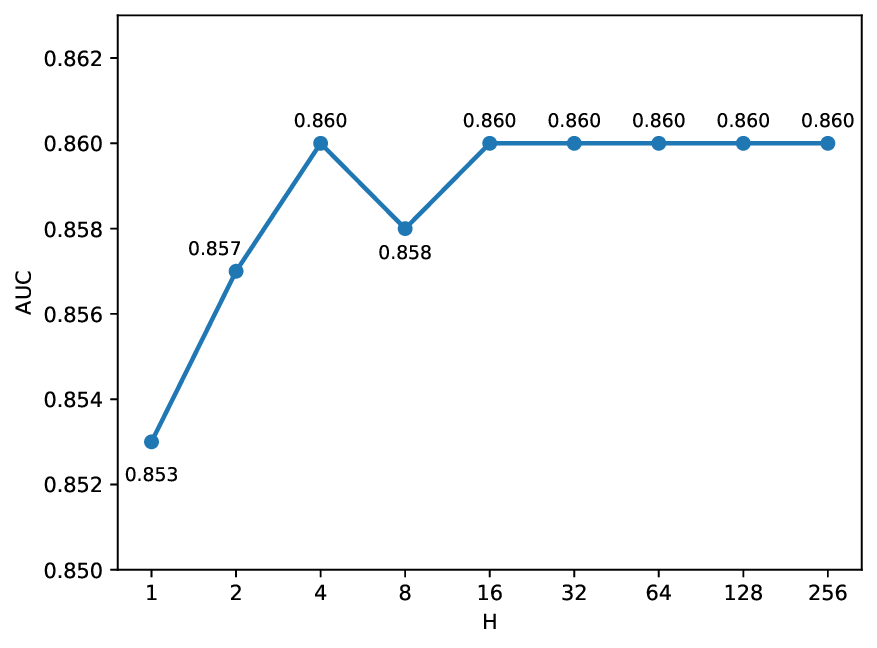} \\ (B)
	\end{minipage}
	
	\caption{Ablation study results for different masking ratios and values of $H$.}
	\label{fig_H_AUC_Mask-Ratio-AUC}
\end{figure*}

Experiments \textbf{e} and \textbf{f} are shown in Figure \ref{fig_H_AUC_Mask-Ratio-AUC}. Specifically:
\begin{itemize}
	\item For \textbf{e}, we evaluate the algorithm under different masking ratios, ranging from 0.5 to 0.95.
	\item For \textbf{f}, we examine the influence of varying the aggregation strategy $H$, with values including 1, 2, 4, 8, 16, 32, 64, 128, and  256 during inference.
\end{itemize}

Figure \ref{fig_H_AUC_Mask-Ratio-AUC}A shows that masking ratios between 0.15 and 0.35 yield optimal performance, which is consistent with previous findings in ECG multi-label classification \citep{zhou2023masked}. Figure \ref{fig_H_AUC_Mask-Ratio-AUC}B illustrates that performance exhibits a slight increase as $H$ grows and stabilizes when $H$ becomes sufficiently large.

\section{Discussion}
\label{sec:dis}
This paper proposes a novel multi-scale masked autoencoder (MAE) framework for anomaly detection in ECG time-series data, achieving state-of-the-art performance on both anomaly detection and localization tasks using the recently released PTB-XL benchmarks \citep{jiang2023multi}. The proposed method models both global and local features within a single, end-to-end Transformer-based architecture. Unlike previous approaches \citep{jiang2023multi, bui2024tsrnet}, our method eliminates the need for heartbeat segmentation or R-peak detection during the data preprocessing stage, which can be inconvenient and unreliable for certain ECG signals in real-world applications. Moreover, the proposed framework is lightweight, with only 0.398M parameters and 0.576 GFLOPs for inference, approximately 1/78 of the FLOPs compared to the previous state-of-the-art method for ECG detection and localization, making it more suitable for clinical deployment compared to existing methods.

A distinctive feature of the proposed approach is the elegant modeling of local features by concatenating subparts of the raw signal and applying distinct positional embeddings, coupled with the multi-scale masking strategy. To the best of our knowledge, this is the first work to integrate these techniques for modeling both global and local features within the MAE framework. We perform ablation studies on the PTB-XL anomaly detection task, confirming the effectiveness of each proposed component. Our experiments demonstrate that multi-scale strategies significantly enhance the MAE's ability to learn more efficient multi-scale features. While previous studies have suggested that MAE might not be optimal for anomaly detection \citep{reiss2022anomaly}, our results indicate that the incorporation of multi-scale strategies substantially improves its performance. On the other hand, it demonstrate that the original MAE architecture, which might focuses on capturing global information, benefits from the inclusion of local feature attention, allowing the model to better capture subtle patterns and details in the input signal.

Furthermore, the masked autoencoder within the Transformer architecture exhibits strong representational learning capabilities for ECG data analysis, extending beyond just anomaly detection. A recent study has demonstrated the application of this architecture in ECG multi-label classification, achieving significant improvements on several benchmark datasets compared to recent state-of-the-art methods \citep{zhou2023masked}. Notably, their approach relies solely on global information. An interesting avenue for future work would be to investigate whether the inclusion of local feature attention can further enhance the performance of ECG multi-label classification models. 

\section{Conclusion}
\label{sec:con}
This paper introduces a novel multi-scale masked autoencoder framework for ECG anomaly detection and localization. By eliminating the reliance on R-peak detection and heartbeat segmentation, 
our approach enhances robustness for real-world clinical ECG recordings. The method employs a novel multi-scale masking strategy with multi-scale attention to capture both global and local dependencies, significantly improving performance while reducing computational complexity to 1/78 of the current state-of-the-art method. Experimental results on the PTB-XL benchmark validate its superior performance and highlight its potential for clinical deployment, particularly in resource-constrained settings. Future work will explore extending this approach to other ECG-related tasks, such as arrhythmia classification and heart disease detection, to further demonstrate its clinical utility.

\section*{Data Availability}
The PTB-XL detection and localization benchmark dataset is publicly available at  
\url{https://github.com/MediaBrain-SJTU/ECGAD}.

\appendix
\section{Appendix}
\subsection{Details of Visualization Examples}
\label{app:detail_exam}
Figure \ref{fig_anomaly_localization} illustrates a portion of ECG signals from different leads of examples in the PTB-XL localization benchmark \citep{jiang2023multi}.
In particular, 
it shows 
the AVR lead from the 212rd sample (A),
the V1 lead from the 225th sample (B), 
the V4 lead from the 230th sample (C),
the V1 lead from the 234th sample (D),
the V2 lead from the 376th sample (E),
and 
the V2 lead from the 389th sample (F). By mapping these samples to the SCP-ECG statements in the PTB-XL dataset \citep{wagner2020ptb}, we identify the corresponding clinical annotations: the 212rd, 225th, 230th, 234th, 376th and 389th samples are labeled with 
[complete left bundle branch block, first degree AV block, premature ventricular contractions],
[complete right bundle branch block,  left posterior fascicular block, first degree AV block, right ventricular hypertrophy],
[left ventricular hypertrophy, non-specific ischemic, incomplete right bundle branch block, first degree AV block],
[inferolateral myocardial infarction,  anteroseptal myocardial infarction],
[anteroseptal myocardial infarction, left ventricular hypertrophy, non-specific ischemic, non-specific intraventricular conduction disturbance (block), left atrial overload/enlargement],
and 
[ischemic in anterolateral leads, ischemic in inferior leads, premature ventricular contractions], respectively.

\bibliographystyle{rss} 
\bibliography{reference}

\end{document}